\documentclass{article}
\usepackage{spconf,amsmath,graphicx}
\usepackage{amssymb,amsfonts}
\usepackage{cite}
\usepackage{url}
\usepackage{booktabs}
\usepackage{multirow}
\usepackage[hidelinks,bookmarks=false]{hyperref}

\begin{document}

\title{ADAPTIVE BAND SELECTION FOR HYPERSPECTRAL CLASSIFICATION WITH SPATIALLY DISJOINT EVALUATION}
\name{Ikram El-Hajri$^{\star}$ \qquad Ouassim Karrakchou$^{\star}$ \qquad Alejandro Mousist$^{\dagger}$%
\thanks{This research has received funding from the European Union's Horizon
research and innovation program under grant agreement No~101070374.}}
\address{$^{\star}$ International University of Rabat, Rabat, Morocco \\
         $^{\dagger}$ Thales Alenia Space, Spain}

\ninept
\maketitle

\setlength{\abovedisplayskip}{4pt}
\setlength{\belowdisplayskip}{4pt}
\setlength{\abovedisplayshortskip}{2pt}
\setlength{\belowdisplayshortskip}{2pt}
\setlength{\textfloatsep}{5pt plus 1pt minus 2pt}
\setlength{\floatsep}{5pt plus 1pt minus 2pt}
\setlength{\intextsep}{5pt plus 1pt minus 2pt}
\setlength{\abovecaptionskip}{2pt}
\setlength{\belowcaptionskip}{1pt}

\begin{abstract}
Hyperspectral band selection methods based on differentiable selectors can be
sensitive to initialization and to extracting a final discrete subset, while
prescribed band counts limit flexibility. We propose SGBR-HC (Spectral-Group Band Ranking with Hard-Concrete
initialization), a two-stage method that uses a supervised spectral
ranking to initialize trainable sparse gates rather than treating ranking
as a fixed selection rule, letting the number of selected bands be
determined by training. Stage-1
scores candidate bands from training pixels by class discriminability and
spectral diversity; this ranking seeds the gate logits for Stage-2, which
trains the sparse gates jointly with a spatial classifier. Under spatially disjoint evaluation on Pavia University and Houston 2013,
verified by retraining a fresh classifier on the selected bands,
SGBR-HC achieves the highest mean overall accuracy and Cohen's $\kappa$ with
approximately twenty bands. Bypassing Stage-1 degrades OA by 8.84~pp on Pavia University and
22.15~pp on Houston~2013, confirming the ranking prior's role.
Random pixel splits inflate OA on Pavia University by 30.56~pp,
underscoring spatial leakage as a critical evaluation confound.
\end{abstract}


\section{Introduction}
\label{sec:intro}

Hyperspectral imaging (HSI) acquires hundreds of narrow, contiguous spectral
channels per scene, producing detailed material signatures that benefit remote
sensing tasks from urban mapping to crop monitoring~\cite{sun2019review}. This spectral richness can worsen the Hughes phenomenon, where classification
performance degrades when the feature dimension grows but labeled training data
remain limited~\cite{hughes1968}. Storage, transmission,
and onboard inference costs also scale with the number of retained channels,
making compact and interpretable spectral representations essential for
practical HSI systems.

Dimensionality reduction for HSI follows two complementary strategies: feature
extraction (FE) and band selection (BS). FE methods compress spectral vectors
into latent components that reduce dimensionality effectively but produce
mixtures of wavelengths and thus lose the original spectral identity; this
complicates physical interpretability and sensor-level deployment. BS selects a
small subset of the original channels, preserving wavelength identity and
informing reduced-band acquisition or omission of specific bands. Consequently,
BS is attractive when interpretability, sensor constraints, or cross-sensor
transfer are priorities. The central BS challenge is to find a compact subset
that retains discriminative power for the downstream task while respecting the
ordered spectral structure.

A large body of BS work computes explicit, interpretable scores or groups from
spectral statistics and selects representatives accordingly. Classical
supervised feature-selection criteria, such as max-dependency/min-redundancy
(mRMR)~\cite{peng2005mrmr}, motivate relevance--redundancy scoring, while
clustering and neighborhood-grouping methods exploit the ordered spectral axis
to reduce local redundancy~\cite{fngbs2021,asps_mn2019,ngnmf2022}.
Grouping arises naturally from spectral contiguity, since adjacent channels are
likely redundant; partitioning the spectrum and selecting representatives
addresses redundancy while preserving wavelength meaning. Self-representation
approaches extend this idea by modeling band dependence explicitly, treating
informative bands as those enabling the reconstruction of others; recent work couples
self-representation with spatial priors to respect spectral--spatial
structure~\cite{tang2023s4p}. These external, criterion-driven methods have the
virtue of explicit rules and interpretability, but because selection precedes classifier training, the chosen subset
optimizes a criterion external to the classification objective used at
evaluation.

To reduce this proxy mismatch, recent research embeds selection into model
training, learning neural selectors jointly with reconstruction or task
objectives. Reconstruction-driven methods form a well-studied branch. BS-Nets~\cite{bsnets2020}
learn band-attention weights jointly with a reconstruction network, Concrete
AE~\cite{sun2022cae} uses a Concrete selector with temperature annealing to learn
a fixed-size subset, stochastic-gate autoencoders~\cite{sun2022sgae} introduce
Gaussian-relaxed sparse gates under a reconstruction objective, and Dropout
CAE~\cite{xu2025dropoutcae} eliminates the post-processing step required to
extract a hard subset by embedding variational dropout directly in the selector. Classification-driven embedded selectors address the reconstruction-proxy
objective by training selectors under task loss~\cite{zimmer2025supervised,shang2025xgbs}.
Nevertheless, both reconstruction- and classification-driven embedded selectors
rely on continuous relaxations, stochastic gates, or soft selection variables
that must eventually be converted into a hard subset. As a result, final
selection can be sensitive to initialization, annealing schedules, and the
conversion from soft to hard subsets, which can destabilize selection and
complicate selection-size control.

Hybrid methods integrate complementary cues, including graphs, dynamic clustering,
reconstruction, classifier feedback, and teacher-guided signals, to better
capture spectral structure and task relevance~\cite{9785994,zhang2026dehf,wu2025s2hgc};
further hybrid approaches combine filter preselection with metaheuristic search,
such as Fisher--ReliefF preprocessing followed by improved dwarf-mongoose
optimization~\cite{lv2025twostage}.
However, to our knowledge, spectral structure in these hybrid methods is used
either as a final fixed selection rule or as part of a learned representation,
rather than as an explicit supervised prior for initializing trainable
sparse-gate parameters. A supervised spectral ranking carries interpretable information about discriminability and
diversity across wavelength regions, while a trainable sparse gate can still
refine the subset during classifier optimization.

The method must therefore preserve wavelength identity and
structured spectral priors, align selection with the classification loss,
stabilize discrete sparse optimization, and avoid prescribing the final number
of bands a priori. A direct way to address these requirements is to bias the
initialization of trainable sparse gates with a supervised spectral ranking,
rather than using ranking as a terminal fixed rule. Hard-Concrete
$\ell_0$-style stochastic gates provide an effective mechanism for this strategy
because they attach stochastic binary gates to model inputs and optimize a
differentiable surrogate for the expected number of active gates~\cite{louizos2018l0}.

In this work, we introduce SGBR-HC (Spectral-Group Band Ranking with
Hard-Concrete initialization), a two-stage hybrid approach. Stage-1 groups bands
along the spectral axis and produces a supervised ranking that combines
class discriminability with inter-group diversity, computed strictly from
training pixels. Stage-2 maps the resulting rank positions into initial logits
for Hard-Concrete stochastic gates that are optimized jointly with a classifier
under an $\ell_0$ regularizer. Without prescribing a band count, the sparsity coefficient $\lambda$
implicitly determines $m$ via the learned gate distribution rather than
requiring a fixed subset size or post-hoc thresholding of soft selection scores.

For reliable evaluation, we apply leakage-aware hard-subset
verification~\cite{sampling2016,ahmad2022disjoint}; Pavia University (PU) uses a spatially disjoint block partition
that ensures no patch crosses partitions, while
Houston~2013 follows the official Data Fusion Contest (DFC) train/test split. We
compare SGBR-HC with published baselines and a random-selection control
under these per-scene evaluation protocols.

The main contributions of this work are:
\begin{itemize}
\item We propose SGBR-HC, a two-stage hybrid selector that uses a supervised
spectral-group ranking to initialize Hard-Concrete sparse-gate logits for
classifier-aligned $\ell_0$ optimization and final hard-subset extraction.

\item We avoid prescribing $m$ directly; instead, the sparsity coefficient
$\lambda$ controls the accuracy--sparsity trade-off, and the realized subset
size is obtained from the learned gate distribution without an additional
post-hoc threshold.

\item Under spatially disjoint evaluation, we quantify a
$30.56$~pp OA gap relative to random pixel splits on PU, showing
that evaluation protocol is a critical confound in BS benchmarks.
\end{itemize}

\section{Method}
\label{sec:method}

Let $\mathbf{X}\in\mathbb{R}^{H\times W\times B}$ denote a hyperspectral image
with $H\times W$ spatial pixels and $B$ spectral bands. The goal is to select
a compact subset of spectral bands that preserves discriminative spectral
information for pixel classification.

\begin{figure*}[t]
\centering
\includegraphics[width=\textwidth]{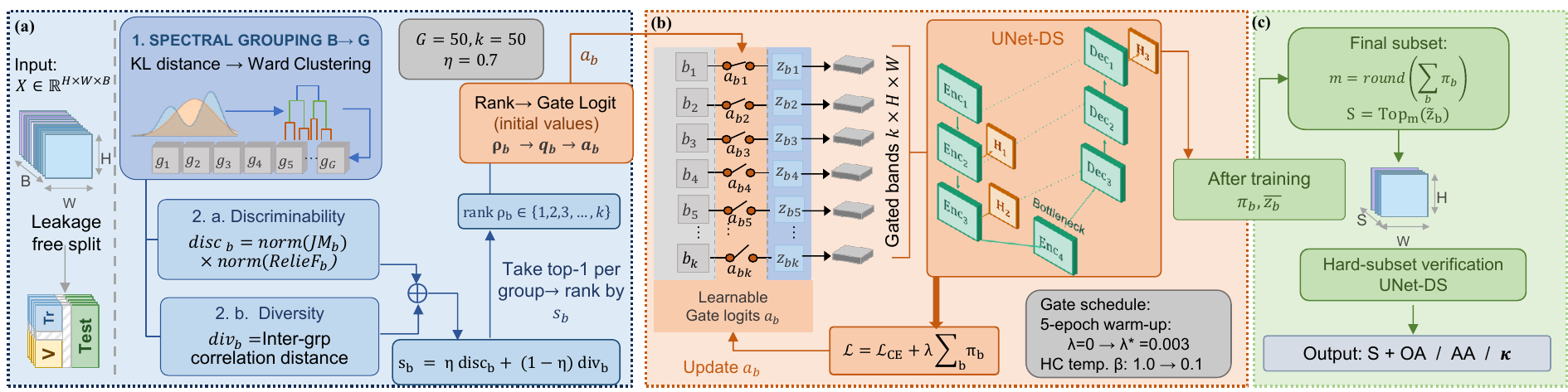}
\caption{Overview of SGBR-HC.
\textbf{(a)} Stage-1 forms spectral groups using KL-based Ward clustering, scores bands by discriminability and inter-group diversity, and produces a ranked top-$k$ candidate pool from train-fold pixels only.
\textbf{(b)} Stage-2 converts the rank position of each candidate into an initial Hard-Concrete gate logit, then jointly trains the gates and a UNet-DS under an $\ell_0$-regularized loss.
\textbf{(c)} The final subset is extracted threshold-free, then verified by training a fresh UNet-DS using only the selected bands.}
\label{fig:pipeline}
\end{figure*}

As illustrated in Fig.~\ref{fig:pipeline}, SGBR-HC performs this selection in
two stages. Stage-1 produces a ranked candidate pool
$\mathcal{C}\subset\{1,\ldots,B\}$ with $|\mathcal{C}|=k$ by grouping bands
according to spectral similarity and scoring them using class discriminability
and inter-group diversity computed from training pixels only. Stage-2 learns
Hard-Concrete gates over the candidates in $\mathcal{C}$, initialized from the
Stage-1 ranks, and optimizes them jointly with the classifier under an $\ell_0$ regularizer. The final subset $\mathcal{S}\subseteq\mathcal{C}$ is then extracted
threshold-free from the learned gates, with $|\mathcal{S}|=m$ determined
by the learned gate distribution rather than fixed in advance.

\subsection{Stage-1: Spectral-Group Band Ranking}

A band's discriminability measures how well its intensity values separate the
target classes; diversity measures how weakly it correlates with bands in other
spectral groups. Ranking by discriminability alone concentrates selections in
one spectral region, because adjacent bands are highly correlated. We therefore
score each band by a weighted combination of both criteria, so the candidate
pool is both class-discriminative and spectrally spread.

\noindent\textbf{Spectral grouping.}
For each band, we estimate a marginal intensity histogram from train-fold
pixels and interpret it as a discrete probability distribution. Band-to-band
dissimilarity is computed with the symmetrized Kullback--Leibler (KL)
divergence between per-band empirical intensity distributions, and
hierarchical clustering with Ward linkage is applied to the resulting
$B\times B$ dissimilarity matrix, following information-measure-based HSI
band-selection clustering~\cite{martinezeuso2007waludi}.
The resulting partition forms $G{=}50$ spectral groups, and $g(b)$ denotes the
group of band~$b$.

\noindent\textbf{Band scoring.}
For each band, we compute a global discriminability score and an inter-group
diversity score. The global discriminability score is
\begin{equation*}
  d_b =
  \mathrm{norm}_{\mathrm{g}}(JM_b)
  \cdot
  \mathrm{norm}_{\mathrm{g}}(\mathrm{ReliefF}_b),
\end{equation*}
where $\mathrm{norm}_{\mathrm{g}}(\cdot)$ denotes global min--max
normalization over all $B$ bands, bringing $JM_b$ and $\mathrm{ReliefF}_b$
to a common scale before multiplication.
$JM_b$ is the sum of Jeffreys-Matusita (JM) divergences
computed for band~$b$ over all class pairs, and
$\mathrm{ReliefF}_b$~\cite{kononenko1994relief} measures local neighborhood
discrimination by rewarding bands that separate nearby different-class samples
while remaining stable among nearby same-class samples.
ReliefF uses up to 500 training pixels and 10 nearest neighbors; JM uses
all Stage-1 training pixels.

The inter-group diversity score is
\begin{equation}
  \delta_b =
  \frac{1}{|\mathcal{I}_b|}
  \sum_{b' \in \mathcal{I}_b}
  \!\left(1 - |c_{bb'}|\right),
  \quad
  \mathcal{I}_b = \{b' : g(b') \neq g(b)\},
  \label{eq:diversity}
\end{equation}
where $c_{bb'}$ is the Pearson correlation between bands $b$ and $b'$
computed on training pixels. Both $d_b$ and $\delta_b$ are then min--max
normalized within each spectral group to make scores comparable across groups,
yielding $d_b^{(g)}$ and $\delta_b^{(g)}$. The final Stage-1 score is
\begin{equation*}
  s_b = \eta\,d_b^{(g)} + (1-\eta)\,\delta_b^{(g)},
\end{equation*}
with $\eta{=}0.7$.

\noindent\textbf{Candidate pool.}
The highest-scoring band from each group is retained as that group's winner,
so no single spectral region fills multiple pool slots. The $G$ winners are
then ranked by their score $s_b$, and the top $k$ form the candidate pool
$\mathcal{C}$ passed to Stage-2. In the final configuration $G{=}k{=}50$,
so all winners enter the pool; their rank still governs the rank-based logit initialization in Stage-2.

\subsection{Stage-2: Hard-Concrete Gating}

Stage-2 assigns a learnable Hard-Concrete gate $z_b$ to each of the $k$
candidates and trains the gates jointly with the classifier under an $\ell_0$ regularizer.

\noindent\textbf{Rank-based initialization.}
The Stage-1 ranking is converted into initial gate logits before optimization.
Let $\rho_b\in\{1,\ldots,k\}$ be the rank of band~$b$ in the Stage-1 list,
with rank~1 assigned to the highest-scoring band. The ranking prior $q_b$,
defined in \eqref{eq:rank_prior}, maps rank to $[0,1]$:
\begin{equation}
  q_b = 1-\frac{\rho_b-1}{k-1}.
  \label{eq:rank_prior}
\end{equation}
The initial gate logit is set by \eqref{eq:gate_init}:
\begin{equation}
  a_b
    = \ell_{\mathrm{low}}+q_b(\ell_{\mathrm{high}}-\ell_{\mathrm{low}}),
  \label{eq:gate_init}
\end{equation}
where $\ell_{\mathrm{low}}{=}\mathrm{logit}(0.02)$ and
$\ell_{\mathrm{high}}{=}\mathrm{logit}(0.90)$, keeping logits finite and avoiding gradient saturation.
All gate logits remain trainable throughout Stage-2.

\noindent\textbf{Gate parameterization.}
Hard-Concrete gating~\cite{louizos2018l0} provides a differentiable
relaxation of binary include/exclude decisions.
During training, each candidate band~$b$ is gated by a stochastic value
$z_b\in[0,1]$ drawn as:
\begin{align*}
  r_b &= \sigma\!\left(\frac{\log u-\log(1-u)+a_b}{\beta}\right),
         \quad u\sim\mathcal{U}(0,1), \\
  z_b &= \min\!\bigl(1,\max(0,r_b(\zeta-\gamma)+\gamma)\bigr),
\end{align*}
where $\sigma(\cdot)$ is the sigmoid function, $u$ is a uniform noise sample
that makes the gate stochastic and the objective differentiable with respect
to $a_b$, $\beta>0$ is a temperature parameter, $\gamma{=}-0.1$, and
$\zeta{=}1.1$. The parameters $\gamma$ and $\zeta$ stretch the support of
$r_b$ beyond $[0,1]$; after clamping, exact zeros and ones occur with positive
probability, so the gate can be truly sparse during training.
The temperature $\beta$ is linearly decayed from $1.0$ to $0.1$ as
training progresses, sharpening the gate distribution from soft to near-binary.

The probability that gate $b$ is nonzero is
\begin{equation*}
  \pi_b = \sigma\!\left(a_b-\beta\log\frac{-\gamma}{\zeta}\right),
\end{equation*}
obtained analytically by marginalizing over $u$. $\pi_b$ is differentiable in
$a_b$ and appears directly in the $\ell_0$ regularizer.

\noindent\textbf{Classifier.}
We use UNet-DS (Deep Supervision), a U-Net-style encoder--decoder with skip connections
adapted for center-pixel hyperspectral classification~\cite{ronneberger2015unet},
as the Stage-2 classifier.
The input is a $17{\times}17$ spatial patch in which each spectral channel is
multiplied by its gate value $z_b$, so channels with $z_b{=}0$ are suppressed.
Two auxiliary classification heads ($H_1$, $H_2$) are attached to intermediate
encoder stages; the main head ($H_3$) is attached to the final decoder stage.
The auxiliary heads improve gradient flow during training; at inference, only
the main head $H_3$ is used.
For all three heads, only the center-pixel logit, written
$(\cdot)^{\mathrm{c}}$, is used for supervision.
The deep-supervision cross-entropy is
\begin{equation*}
  \mathcal{L}_{\mathrm{CE}}
  = 0.2\,\mathrm{CE}(o_1^{\mathrm{c}},y)
  + 0.3\,\mathrm{CE}(o_2^{\mathrm{c}},y)
  + 0.5\,\mathrm{CE}(o_3^{\mathrm{c}},y),
\end{equation*}
where $(o_1,o_2,o_3)$ are the logit maps from the three heads and $y$ is the
ground-truth class label of the center pixel.

\noindent\textbf{Training objective.}
The gates and classifier are trained jointly by minimizing the objective in
\eqref{eq:l0_loss}:
\begin{equation}
  \mathcal{L}=\mathcal{L}_{\mathrm{CE}}+
  \lambda\sum_{b=1}^{k}\pi_b,
  \label{eq:l0_loss}
\end{equation}
where $\lambda\sum_{b}\pi_b$ penalizes the expected number of open gates.
Both the temperature decay and the $\lambda$ penalty are delayed by a 5-epoch
warm-up, during which gates remain soft and unpenalized so the classifier
first becomes informative before gating decisions sharpen and are regularized.

\noindent\textbf{Final subset extraction.}
At inference, each gate is replaced by its deterministic value:
\begin{equation*}
  \bar{z}_b = \min\!\bigl(1,\max(0,\sigma(a_b)(\zeta-\gamma)+\gamma)\bigr).
\end{equation*}
Rather than applying a fixed threshold on $\bar{z}_b$, which would require
choosing an additional hyperparameter, the realized band count is set by
\eqref{eq:m_count}:
\begin{equation}
  m = \mathrm{round}\!\left(\sum_{b=1}^{k} \pi_b\right).
  \label{eq:m_count}
\end{equation}
The $m$ candidates with the largest $\bar{z}_b$ values form the
selected subset $\mathcal{S}$.

\section{Experiments}
\label{sec:experiments}

\subsection{Datasets and Evaluation Protocol}

\textbf{PU~\cite{ehu_pavia}.}
The ROSIS-03 PU scene contains $610\times340$ pixels, 103 usable bands, and 9 classes.
Following the leakage-free block partition (LFBP) of Jia and
Ding~\cite{app16073543}, the image is partitioned into non-overlapping
$B_s{\times}B_s$ blocks, each assigned to train, validation, or test; a buffer
of $h{=}8$ pixels (equal to the patch radius $r{=}(p{-}1)/2$ for patch size
$p{=}17$, footprint $17{\times}17$) is excluded at every boundary so that all
extracted patches lie fully within their assigned block.
We use $B_s{=}32$; the block-to-partition assignment is randomly permuted across seeds.
A class-aware repair step (our extension) reduces the effective block size
(i.e., subdivides blocks) when buffer filtering leaves a class uncovered,
yielding effective block sizes of 22--32 pixels.

\textbf{Houston 2013~\cite{debes2014houston}.}
The DFC~2013 Houston scene contains $349\times1905$ pixels, 144 bands, and
15 classes. Unlike PU, Houston~2013 provides an official contest train/test
partition, with $2{,}832$ labeled training pixels and $12{,}197$ labeled test
pixels. We retain this official split to preserve comparability with the
standard benchmark, and draw a seed-specific validation subset from the
official training mask. The remaining official training pixels are used for
training, while the official test mask is never used for model selection.
UNet-DS is evaluated on $17{\times}17$ center-pixel patches using the same
patch size and training procedure as PU.

\subsection{Implementation Details}

\begin{figure}[!t]
    \centering
    \includegraphics[width=\columnwidth]{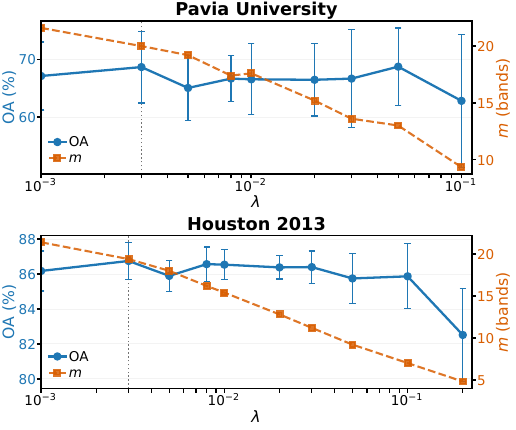}
    \caption{Validation OA (\%) and selected band count $m$ vs.\ $\lambda$ on PU (top) and
    Houston~2013 (bottom); each point is the mean $\pm$ std over five seeds and the dotted
    line marks $\lambda^*{=}0.003$. OA remains stable across $\lambda\in[0.001,0.01]$ before
    degrading under stronger sparsity as $m$ continues to shrink.}
    \label{fig:lambda}
\end{figure}

Stage-1 uses $G{=}50$ spectral groups and $\eta{=}0.7$.
The candidate pool size $k$ is chosen from $\{25,50,B\}$: $k{=}25$ under-supplies Stage-2,
while $k{=}B$ removes the candidate-pool reduction step; both alternatives
reduce OA on both datasets, so we fix $k{=}50$.
The sparsity coefficient $\lambda$ is selected by a 10-point logarithmic
grid search over $\lambda\in[0.001,0.2]$; both datasets select
$\lambda^*{=}0.003$, with OA stable across $\lambda\in[0.001,0.01]$ before
degrading under stronger sparsity (Fig.~\ref{fig:lambda}).
Temperature $\beta$ is linearly decayed from $1.0$ to $0.1$ after a 5-epoch warm-up.
All UNet-DS training uses Adam ($\mathrm{lr}{=}10^{-3}$, batch size~32) with
StepLR decay (factor $0.1$, step 10 epochs).
Stage-2 trains for 40 epochs with inverse-class-frequency sample weights to
compensate for class imbalance, followed by up to 20 fine-tuning epochs with
early stopping (patience~10).

\subsection{Baselines and Metrics}

\textit{Random-$m$} draws a deterministic uniform sample of $m$ bands
without replacement, seeded per run using that seed's SGBR-HC realized band count.
Dropout CAE~\cite{xu2025dropoutcae} is trained end-to-end for a fixed
target band count. CHBS~\cite{zimmer2025supervised} uses the authors'
official architecture, trained on the training fold only.
DEHF~\cite{zhang2026dehf} uses the authors' released code.
NGNMF~\cite{ngnmf2022} uses only the BS stage, following the
original paper. BS-Nets~\cite{bsnets2020}, FNGBS~\cite{fngbs2021},
ASPS-MN~\cite{asps_mn2019}, S$^4$P~\cite{tang2023s4p}, and
mRMR~\cite{peng2005mrmr} use publicly released implementations.

For all BS methods, $m$ is matched to SGBR-HC's realized count
on the same seed ($20.0{\pm}1.3$ bands on PU; $19.4{\pm}0.5$ on Houston~2013). We term this hard-subset verification, where selected subsets are frozen and
evaluated by retraining a fresh UNet-DS on $17{\times}17$ center-pixel
patches with dropout~$0.1$, using the training split only and validation OA
for early stopping. We report
OA, average accuracy (AA), and Cohen's $\kappa$ on the
test split.

\subsection{Main Results}
\label{sec:main_results}

Tables~\ref{tab:main_pavia} and~\ref{tab:main_houston} compare SGBR-HC against
nine published BS methods and a random-selection control.

\begin{table}[t]
\centering
\caption{PU ($B_s{=}32$, $h{=}8$, 5 seeds, retrained $17{\times}17$-patch UNet-DS). All BS methods use $m{=}20.0{\pm}1.3$ bands matched to SGBR-HC per seed. \textbf{Bold}: best.}
\label{tab:main_pavia}
\vspace{2pt}
\renewcommand{\arraystretch}{1.05}
\setlength{\tabcolsep}{4pt}
\footnotesize
\resizebox{\columnwidth}{!}{%
\begin{tabular}{lrrr}
\toprule
Method & OA (\%) & AA (\%) & $\kappa$ \\
\midrule
All bands (103) & $67.11{\pm}7.13$ & $63.94{\pm}5.67$ & $0.566{\pm}0.075$ \\
\midrule
ASPS-MN~\cite{asps_mn2019}          & $68.17{\pm}4.16$ & $\mathbf{66.57{\pm}5.61}$ & $0.580{\pm}0.040$ \\
BS-Nets~\cite{bsnets2020}           & $68.50{\pm}4.97$ & $65.48{\pm}5.77$ & $0.583{\pm}0.053$ \\
CHBS~\cite{zimmer2025supervised}    & $68.08{\pm}6.73$ & $66.02{\pm}6.93$ & $0.579{\pm}0.075$ \\
DEHF~\cite{zhang2026dehf}           & $64.61{\pm}6.55$ & $64.64{\pm}5.35$ & $0.543{\pm}0.058$ \\
Dropout CAE~\cite{xu2025dropoutcae} & $64.73{\pm}9.49$ & $63.17{\pm}5.13$ & $0.540{\pm}0.096$ \\
FNGBS~\cite{fngbs2021}              & $66.03{\pm}11.20$ & $65.01{\pm}5.48$ & $0.562{\pm}0.114$ \\
mRMR~\cite{peng2005mrmr}            & $65.07{\pm}8.10$ & $63.51{\pm}3.60$ & $0.546{\pm}0.072$ \\
NGNMF~\cite{ngnmf2022}              & $67.97{\pm}8.36$ & $66.12{\pm}4.76$ & $0.582{\pm}0.084$ \\
S$^4$P~\cite{tang2023s4p}           & $65.75{\pm}7.09$ & $66.11{\pm}6.59$ & $0.556{\pm}0.074$ \\
Random-$m$                          & $67.60{\pm}4.86$ & $65.48{\pm}4.67$ & $0.574{\pm}0.049$ \\
\midrule
\textbf{Ours (SGBR-HC)}             & $\mathbf{69.35{\pm}6.55}$ & $66.04{\pm}5.23$ & $\mathbf{0.594{\pm}0.070}$ \\
\bottomrule
\end{tabular}
}
\end{table}

\begin{table}[t]
\centering
\caption{Houston 2013 (official DFC~2013 split, 5 seeds, retrained $17{\times}17$-patch UNet-DS). All BS methods use $m{=}19.4{\pm}0.5$ bands matched to SGBR-HC per seed. \textbf{Bold}: best.}
\label{tab:main_houston}
\vspace{2pt}
\renewcommand{\arraystretch}{1.05}
\setlength{\tabcolsep}{4pt}
\footnotesize
\resizebox{\columnwidth}{!}{%
\begin{tabular}{lrrr}
\toprule
Method & OA (\%) & AA (\%) & $\kappa$ \\
\midrule
All bands (144) & $84.94{\pm}1.33$ & $84.42{\pm}0.99$ & $0.836{\pm}0.015$ \\
\midrule
ASPS-MN~\cite{asps_mn2019}          & $85.55{\pm}0.91$ & $86.40{\pm}1.13$ & $0.843{\pm}0.010$ \\
BS-Nets~\cite{bsnets2020}           & $85.54{\pm}0.42$ & $85.72{\pm}0.27$ & $0.843{\pm}0.004$ \\
CHBS~\cite{zimmer2025supervised}    & $84.92{\pm}1.38$ & $85.79{\pm}0.88$ & $0.836{\pm}0.015$ \\
DEHF~\cite{zhang2026dehf}           & $84.66{\pm}0.83$ & $85.24{\pm}0.83$ & $0.833{\pm}0.009$ \\
Dropout CAE~\cite{xu2025dropoutcae} & $86.08{\pm}0.74$ & $\mathbf{86.47{\pm}0.85}$ & $0.849{\pm}0.008$ \\
FNGBS~\cite{fngbs2021}              & $85.78{\pm}1.25$ & $86.40{\pm}0.94$ & $0.845{\pm}0.014$ \\
mRMR~\cite{peng2005mrmr}            & $85.69{\pm}0.69$ & $85.56{\pm}0.75$ & $0.845{\pm}0.007$ \\
NGNMF~\cite{ngnmf2022}              & $84.71{\pm}0.80$ & $85.05{\pm}1.11$ & $0.834{\pm}0.009$ \\
S$^4$P~\cite{tang2023s4p}           & $82.56{\pm}2.61$ & $82.18{\pm}2.92$ & $0.811{\pm}0.028$ \\
Random-$m$                          & $85.65{\pm}0.67$ & $86.00{\pm}1.00$ & $0.844{\pm}0.007$ \\
\midrule
\textbf{Ours (SGBR-HC)}             & $\mathbf{86.24{\pm}1.61}$ & $86.13{\pm}1.61$ & $\mathbf{0.851{\pm}0.018}$ \\
\bottomrule
\end{tabular}
}
\end{table}

On PU, SGBR-HC achieves the highest mean OA ($69.35{\pm}6.55\%$) and $\kappa$,
leading the next-best method (BS-Nets, $68.50\%$) by $0.85$~pp; SGBR-HC also
exceeds the All-bands baseline ($67.11\%$), confirming that ${\sim}20$ bands are
sufficient to surpass using all 103.
SGBR-HC ranks first in OA across all five seeds, confirming the result is
not driven by a single favorable partition.
Several embedded selectors relying on continuous relaxations, including
Dropout CAE ($64.73{\pm}9.49\%$) and DEHF ($64.61{\pm}6.55\%$), fall below
All-bands on PU; the elevated Dropout CAE variance suggests that soft-to-hard
conversion instability is amplified by the variable training-set composition
of the block+buffer protocol.
On Houston~2013, SGBR-HC achieves the highest mean OA ($86.24{\pm}1.61\%$)
and $\kappa$, leading the next-best method (Dropout CAE, $86.08\%$) by $0.16$~pp;
the lower seed variance relative to PU reflects the fixed DFC~2013 test set,
in contrast to PU where both training and test composition vary across seeds.
SGBR-HC trails the leading AA method by $0.53$~pp on PU (ASPS-MN, $66.57\%$)
and $0.34$~pp on Houston (Dropout CAE, $86.47\%$); the $\ell_0$ regularizer
compresses the subset without explicit class-specific band weighting, which
may sacrifice wavelengths that primarily benefit minority classes despite
inverse-frequency training weights.
Across both datasets, SGBR-HC is the only method to rank first on both OA
and $\kappa$, achieving this at ${\sim}20$ bands.

\subsection{Ablation Study}
\label{sec:ablations}

Table~\ref{tab:abl_stage1} evaluates the contribution of the Stage-1 spectral
prior and the Stage-2 sparse-gate refinement.
\textit{SGBR Stage-1 only} evaluates all 50 Stage-1 candidates without
Hard-Concrete refinement.
\textit{Shuffled init} keeps the same 50 Stage-1 candidate bands but randomly
permutes $\rho_b$ (Eq.~\ref{eq:rank_prior}) before logit initialization
(Eq.~\ref{eq:gate_init}).
\textit{Uniform init} bypasses Stage-1 entirely
(Eqs.~\ref{eq:rank_prior}--\ref{eq:gate_init}) and initializes all gates
uniformly over all input bands.
\textit{No diversity} ($\eta{=}1$, removing $\delta_b$ in Eq.~\ref{eq:diversity})
retains Ward clustering and JM${\times}$ReliefF but removes the diversity term.
\textit{JM-only ranking} ranks bands using JM divergence alone, removing both
Ward grouping and the diversity term (Eq.~\ref{eq:diversity}).

\begin{table}[t]
\centering
\caption{Stage-1 design ablation on PU and Houston~2013 ($\lambda^*{=}0.003$, 5~seeds, hard-subset UNet-DS). \textbf{Bold}: best.}
\label{tab:abl_stage1}
\vspace{2pt}
\renewcommand{\arraystretch}{1.05}
\setlength{\tabcolsep}{4pt}
\footnotesize
\resizebox{\columnwidth}{!}{%
\begin{tabular}{lrrr}
\toprule
\multicolumn{4}{c}{\textit{PU}} \\
\midrule
Method & OA (\%) & AA (\%) & $\kappa$ \\
\midrule
\textbf{Ours (SGBR-HC)} & $\mathbf{69.35{\pm}6.55}$ & $66.04{\pm}5.23$ & $\mathbf{0.594{\pm}0.070}$ \\
SGBR Stage-1 only        & $67.69{\pm}4.71$ & $64.11{\pm}5.36$ & $0.569{\pm}0.057$ \\
Shuffled init            & $66.84{\pm}7.08$ & $\mathbf{66.36{\pm}6.06}$ & $0.569{\pm}0.075$ \\
Uniform init (no Stage-1)& $60.51{\pm}7.94$ & $49.16{\pm}6.45$ & $0.470{\pm}0.084$ \\
No diversity ($\eta{=}1$) & $66.64{\pm}7.23$ & $63.53{\pm}6.81$ & $0.563{\pm}0.078$ \\
JM-only ranking          & $69.22{\pm}5.35$ & $61.51{\pm}9.05$ & $0.581{\pm}0.069$ \\
\midrule
\multicolumn{4}{c}{\textit{Houston~2013}} \\
\midrule
\textbf{Ours (SGBR-HC)} & $\mathbf{86.24{\pm}1.61}$ & $\mathbf{86.13{\pm}1.61}$ & $\mathbf{0.851{\pm}0.018}$ \\
SGBR Stage-1 only        & $84.39{\pm}0.59$ & $84.17{\pm}0.69$ & $0.830{\pm}0.006$ \\
Shuffled init            & $84.99{\pm}2.10$ & $85.77{\pm}2.28$ & $0.837{\pm}0.023$ \\
Uniform init (no Stage-1)& $64.09{\pm}0.60$ & $64.74{\pm}0.67$ & $0.610{\pm}0.006$ \\
No diversity ($\eta{=}1$) & $78.04{\pm}1.04$ & $77.57{\pm}1.22$ & $0.762{\pm}0.011$ \\
JM-only ranking          & $79.18{\pm}0.45$ & $78.16{\pm}0.58$ & $0.774{\pm}0.005$ \\
\bottomrule
\end{tabular}
}
\end{table}

\textit{SGBR Stage-1 only} establishes Stage-2's baseline contribution:
Hard-Concrete gating improves OA by $1.66$~pp on PU and $1.85$~pp on
Houston~2013 while compressing the subset from 50 to ${\sim}20$ bands.
Rank-based initialization contributes a useful but secondary signal: retaining
the Stage-1 pool while randomizing logit order (\textit{Shuffled init}) reduces
OA by $2.51$~pp on PU and $1.25$~pp on Houston~2013.
\textit{Shuffled init} posts the highest AA on PU ($66.36\%$), marginally
above SGBR-HC ($66.04\%$); without rank-based initialization, Stage-2 selects
a more spectrally spread subset that benefits minority classes at the cost of
$2.51$~pp in OA.
In contrast, \textit{Uniform init} causes much larger drops of $8.84$~pp on PU
and $22.15$~pp on Houston~2013; near-zero Jaccard overlap between selected band
sets ($J{=}|S{\cap}S'|/|S{\cup}S'|$; $J{=}0.083{\pm}0.056$ on PU;
$J{=}0.037{\pm}0.013$ on Houston) confirms that $\ell_0$ optimization without
the Stage-1 prior converges to a qualitatively different subset.
Since this variant removes both the Stage-1 candidate pool and the rank-based
initialization, subtracting the Shuffled-init drop isolates the pool contribution
at $6.33$~pp on PU and $20.90$~pp on Houston, confirming the candidate pool as
the dominant factor.

The Stage-1 scoring ablations further show the importance of spectral diversity.
On Houston~2013, removing the diversity term reduces OA by $8.20$~pp, while
\textit{JM-only ranking} reduces OA by $7.06$~pp.
These drops are larger than on PU, suggesting that diversity preservation is
more consequential for the 15-class Houston scene, where fine-grained urban
classes require wavelengths from distinct spectral regions that a
discriminability-only ranking tends to concentrate in a single cluster.
On PU, \textit{JM-only ranking} nearly matches SGBR-HC in OA but trails by
$4.53$~pp in AA, indicating that grouping and ReliefF may help preserve
minority-class performance that OA alone can obscure.

\subsection{Effect of Evaluation Protocol}
\label{sec:ablation_protocol}

Random pixel assignment, the most common HSI evaluation practice, enforces
only pixel-level disjointness, but under
strong spatial autocorrelation~\cite{sampling2016} adjacent pixels share
nearly identical spectra, so a test pixel neighboring a training pixel is
not an independent sample.
The index-level partition of Ahmad et al.~\cite{ahmad2022disjoint} goes
further by enforcing strict geographic disjointness
($\mathrm{Train}\cap\mathrm{Val}\cap\mathrm{Test}=\emptyset$) at the pixel
level, ensuring that training, validation, and test samples come from separate
spatial locations, yet applies no patch-radius buffer; center pixels from
different splits that happen to be adjacent still produce overlapping patches,
so geographically disjoint $\neq$ spatially independent for patch-based
classifiers.

Following Jia and Ding's LFBP~\cite{app16073543}, we adopt block+buffer
evaluation, which addresses both leakage sources by assigning whole spatial
blocks to one partition and excluding a patch-radius buffer at every boundary.
On PU, the $36.91$~pp OA gap between the Ahmad et al.\ split and block+buffer
on a per-pixel MLP (no receptive field) shows that autocorrelation-driven
leakage is not limited to patch-based classifiers.
A separate comparison on the UNet-DS patch classifier using SGBR-HC bands
shows a $30.56$~pp gap between random pixel assignment and block+buffer,
confirming that patch overlap compounds the effect.
The two experiments involve different classifiers and split protocols and
are not directly comparable in magnitude, but both motivate block+buffer
as the evaluation standard throughout this work.

\section{Conclusion}
\label{sec:conclusion}

We presented SGBR-HC, a two-stage hyperspectral BS method that
uses a supervised spectral-group ranking to initialize
Hard-Concrete sparse-gate logits.
Under leakage-aware hard-subset verification, SGBR-HC achieves the
highest mean OA and $\kappa$ on PU and ranks first on both OA and $\kappa$
on Houston~2013 by a narrow margin.
Compared with \textit{SGBR Stage-1 only}, SGBR-HC improves OA by
$1.66$~pp on PU and $1.85$~pp on Houston while selecting about 20 bands.
Stage-1 is the dominant factor; bypassing it causes up to $22.15$~pp OA
degradation with near-zero subset overlap, confirming that $\ell_0$ optimization is sensitive to the candidate pool
supplied by Stage-1; within-pool rank ordering provides a secondary but
useful signal (Shuffled init: $-2.51$~pp PU, $-1.25$~pp Houston).
We further show that random pixel splits inflate OA by $30.56$~pp on PU
relative to spatially disjoint evaluation, underscoring that evaluation protocol
is a critical confound in BS benchmarks.
Both evaluation datasets are urban airborne scenes, and hard-subset
verification relies on a single UNet-DS classifier architecture; future
work should test broader scene types, sensor modalities, and classifier
families to assess generalization.

\bibliographystyle{IEEEbib}
\bibliography{references}

\begin{thebibliography}{10}

\bibitem{sun2019review}
Weiwei Sun and Qian Du,
\newblock ``Hyperspectral band selection: A review,''
\newblock {\em IEEE Geoscience and Remote Sensing Magazine}, vol. 7, no. 2, pp.
  118--139, 2019.

\bibitem{hughes1968}
G.~F. Hughes,
\newblock ``On the mean accuracy of statistical pattern recognizers,''
\newblock {\em IEEE Transactions on Information Theory}, vol. 14, no. 1, pp.
  55--63, 1968.

\bibitem{peng2005mrmr}
Hanchuan Peng, Fuhui Long, and Chris Ding,
\newblock ``Feature selection based on mutual information: Criteria of
  max-dependency, max-relevance, and min-redundancy,''
\newblock {\em IEEE Transactions on Pattern Analysis and Machine Intelligence},
  vol. 27, no. 8, pp. 1226--1238, 2005.

\bibitem{fngbs2021}
Qi~Wang, Qiang Li, and Xuelong Li,
\newblock ``A fast neighborhood grouping method for hyperspectral band
  selection,''
\newblock {\em IEEE Transactions on Geoscience and Remote Sensing}, vol. 59,
  no. 6, pp. 5028--5039, 2021.

\bibitem{asps_mn2019}
Qi~Wang, Qiang Li, and Xuelong Li,
\newblock ``Hyperspectral band selection via adaptive subspace partition
  strategy,''
\newblock {\em IEEE Journal of Selected Topics in Applied Earth Observations
  and Remote Sensing}, vol. 12, no. 12, pp. 4940--4950, 2019.

\bibitem{ngnmf2022}
Hang Fu, Aizhu Zhang, Genyun Sun, Jinchang Ren, Xiuping Jia, Zhaojie Pan, and
  Hongzhang Ma,
\newblock ``A novel band selection and spatial noise reduction method for
  hyperspectral image classification,''
\newblock {\em IEEE Transactions on Geoscience and Remote Sensing}, vol. 60,
  pp. 1--13, 2022.

\bibitem{tang2023s4p}
Chang Tang, Jun Wang, Xiao Zheng, Xinwang Liu, Weiying Xie, Xianju Li, and
  Xinzhong Zhu,
\newblock ``Spatial and spectral structure preserved self-representation for
  unsupervised hyperspectral band selection,''
\newblock {\em IEEE Transactions on Geoscience and Remote Sensing}, vol. 61,
  pp. 1--13, 2023.

\bibitem{bsnets2020}
Yaoming Cai, Xiaobo Liu, and Zhihua Cai,
\newblock ``{BS-Nets}: An end-to-end framework for band selection of
  hyperspectral image,''
\newblock {\em IEEE Transactions on Geoscience and Remote Sensing}, vol. 58,
  no. 3, pp. 1969--1984, 2020.

\bibitem{sun2022cae}
He~Sun, Jinchang Ren, Huimin Zhao, Peter Yuen, and Julius Tschannerl,
\newblock ``Novel {Gumbel-Softmax} trick enabled concrete autoencoder with
  entropy constraints for unsupervised hyperspectral band selection,''
\newblock {\em IEEE Transactions on Geoscience and Remote Sensing}, vol. 60,
  pp. 1--13, 2022.

\bibitem{sun2022sgae}
He~Sun, Lei Zhang, Lizhi Wang, and Hua Huang,
\newblock ``Stochastic gate-based autoencoder for unsupervised hyperspectral
  band selection,''
\newblock {\em Pattern Recognition}, vol. 132, pp. 108969, 2022.

\bibitem{xu2025dropoutcae}
Lei Xu, Mete Ahishali, and Moncef Gabbouj,
\newblock ``Dropout concrete autoencoder for band selection on hyperspectral
  image scenes,''
\newblock {\em IEEE Geoscience and Remote Sensing Letters}, vol. 22, pp. 1--5,
  2025.

\bibitem{zimmer2025supervised}
Yaniv Zimmer, Ofir Lindenbaum, and Oren Glickman,
\newblock ``Supervised embedded methods for hyperspectral band selection,''
\newblock in {\em ECAI 2025 -- 28th European Conference on Artificial
  Intelligence, including 14th Conference on Prestigious Applications of
  Intelligent Systems, PAIS 2025 -- Proceedings}. Oct. 2025, vol. 413 of {\em
  Frontiers in Artificial Intelligence and Applications}, pp. 3154--3161, IOS
  Press.

\bibitem{shang2025xgbs}
Xiaodi Shang, Chuanyu Cui, Xudong Sun, Xiaopeng Wang, and Jiahua Zhang,
\newblock ``Classification task-driven hyperspectral band selection via
  interpretability from {XGBoost},''
\newblock {\em IEEE Journal of Selected Topics in Applied Earth Observations
  and Remote Sensing}, vol. 18, pp. 13733--13754, 2025.

\bibitem{9785994}
Chunyan Yu, Sijia Zhou, Meiping Song, Baoyu Gong, Enyu Zhao, and Chein-I Chang,
\newblock ``Unsupervised hyperspectral band selection via hybrid graph
  convolutional network,''
\newblock {\em IEEE Transactions on Geoscience and Remote Sensing}, vol. 60,
  pp. 1--15, 2022.

\bibitem{zhang2026dehf}
Jiahua Zhang, Jingyuan Wang, Dejie Li, Jun Zhang, Xudong Sun, and Xiaodi Shang,
\newblock ``Hyperspectral band selection with dynamic graph enhancement and
  hierarchical feature fusion,''
\newblock {\em IEEE Transactions on Geoscience and Remote Sensing}, vol. 64,
  pp. 1--16, 2026.

\bibitem{wu2025s2hgc}
Kaixiong Wu, Mingwei Wang, Haipeng Luo, Maolin Chen, Yujie Xu, and Wei Liu,
\newblock ``{S$^2$HGC}: An end-to-end spectral--spatial hypergraph
  convolutional network for unsupervised hyperspectral band selection,''
\newblock {\em IEEE Transactions on Geoscience and Remote Sensing}, vol. 63,
  pp. 1--16, 2025.

\bibitem{lv2025twostage}
Huanhuan Lv, Chong Chen, Hui Zhang, Cuiping Shi, and Ruiqin Wang,
\newblock ``A two-stage feature selection method for hyperspectral image
  classification: Hybrid filter preprocessing and improved dwarf mongoose
  optimization,''
\newblock {\em IEEE Journal of Selected Topics in Applied Earth Observations
  and Remote Sensing}, vol. 18, pp. 29271--29294, 2025.

\bibitem{louizos2018l0}
Christos Louizos, Max Welling, and Diederik~P. Kingma,
\newblock ``Learning sparse neural networks through {$L_0$} regularization,''
\newblock in {\em International Conference on Learning Representations}, 2018.

\bibitem{sampling2016}
Jie Liang, Jun Zhou, Yuntao Qian, Lian Wen, Xiao Bai, and Yongsheng Gao,
\newblock ``On the sampling strategy for evaluation of spectral-spatial methods
  in hyperspectral image classification,''
\newblock {\em IEEE Transactions on Geoscience and Remote Sensing}, vol. 55,
  no. 2, pp. 862--880, 2017.

\bibitem{ahmad2022disjoint}
Muhammad Ahmad, Usman Ghous, Danfeng Hong, Adil~Mehmood Khan, Jing Yao, Shaohua
  Wang, and Jocelyn Chanussot,
\newblock ``A disjoint samples-based {3D-CNN} with active transfer learning for
  hyperspectral image classification,''
\newblock {\em IEEE Transactions on Geoscience and Remote Sensing}, vol. 60,
  pp. 1--16, 2022.

\bibitem{martinezeuso2007waludi}
Adolfo Mart{\'i}nez-Us{\'o}, Filiberto Pla, Jos{\'e} Mart{\'i}nez~Sotoca, and
  Pedro Garc{\'i}a-Sevilla,
\newblock ``Clustering-based hyperspectral band selection using information
  measures,''
\newblock {\em IEEE Transactions on Geoscience and Remote Sensing}, vol. 45,
  no. 12, pp. 4158--4171, 2007.

\bibitem{kononenko1994relief}
Igor Kononenko,
\newblock ``Estimating attributes: Analysis and extensions of {RELIEF},''
\newblock in {\em Machine Learning: {ECML}-94}. 1994, pp. 171--182, Springer.

\bibitem{ronneberger2015unet}
Olaf Ronneberger, Philipp Fischer, and Thomas Brox,
\newblock ``U-net: Convolutional networks for biomedical image segmentation,''
\newblock in {\em Medical Image Computing and Computer-Assisted Intervention --
  MICCAI 2015}. 2015, pp. 234--241, Springer.

\bibitem{ehu_pavia}
{Grupo de Inteligencia Computacional, Universidad del Pa{\'i}s Vasco},
\newblock ``Hyperspectral remote sensing scenes,''
  \url{https://www.ehu.eus/ccwintco/index.php/Hyperspectral_Remote_Sensing_Scenes#Pavia_University},
\newblock Accessed: 2026-03-10.

\bibitem{app16073543}
Tong Jia and Haiyong Ding,
\newblock ``Leakage-free evaluation and multi-prototype contrastive learning
  for hyperspectral classification of vegetation,''
\newblock {\em Applied Sciences}, vol. 16, no. 7, 2026.

\bibitem{debes2014houston}
Christian Debes, Andreas Merentitis, Roel Heremans, J{\"u}rgen Hahn, Nikolaos
  Frangiadakis, Tim van Kasteren, Wenzhi Liao, Rik Bellens, Aleksandra
  Pi{\v{z}}urica, Sidharta Gautama, Wilfried Philips, Saurabh Prasad, Qian Du,
  and Fabio Pacifici,
\newblock ``Hyperspectral and {LiDAR} data fusion: Outcome of the 2013 {GRSS}
  data fusion contest,''
\newblock {\em IEEE Journal of Selected Topics in Applied Earth Observations
  and Remote Sensing}, vol. 7, no. 6, pp. 2405--2418, 2014.

\end{thebibliography}

\end{document}